# A Survey on Dragonfly Algorithm and its Applications in Engineering

Chnoor M. Rahman[1], Tarik A. Rashid[2*], Abeer Alsadoon[3,4,5,6], Nebojsa Bacanin[7], Polla Fattah[8] , Seyedali Mirjalili [9,10]

[1]*Applied Computer Department, College of Health and Applied Sciences, Charmo University, Sulaimany, Iraq, chnoor.rahman@charmouniversity.org*
[2*] *Science and Engineering Department, University of Kurdistan Hewler, Erbil Iraq, tarik.ahmed@ukh.edu.krd*
[3] *School of Computing and Mathematics, Charles Sturt University, Sydney, Australia*
[4]*School of Computing Engineering and Mathematics, Western Sydney University, Sydney City Campus, Australia*
[5]*Asia Pacific International College (APIC), Information Technology Department, Sydney, Australia*
[6]*Kent Institute Australia, Information Technology Department, Sydney, Australia, alsadoon.abeer@gmail.com*
[7] *Singidunum University, Belgrade, Serbia, nbacanin@singidunum.ac.rs*
[8]*Software and Informatics Engineering, Salahaddin University-Erbil-, Erbil, Iraq, pollaeng@gmail.com*
[9]*Centre for Artificial Intelligence Research and Optimisation, Torrens University, Australia*
[10]*Yonsei Frontier Lab, Yonsei University, Seoul, Korea; ali.mirjalili@gmail.com*

## Abstract

The dragonfly algorithm was developed in 2016. It is one of the algorithms used by researchers to optimize an extensive series of uses and applications in various areas. At times, it offers superior performance compared to the most well-known optimization techniques. However, this algorithm faces several difficulties when it is utilized to enhance complex optimization problems. This work addressed the robustness of the method to solve real-world optimization issues, and its deficiency to improve complex optimization problems. This review paper shows a comprehensive investigation of the dragonfly algorithm in the engineering area. First, an overview of the algorithm is discussed. Besides, we also examined the modifications of the algorithm. The merged forms of this algorithm with different techniques and the modifications that have been done to make the algorithm perform better are addressed. Additionally, a survey on applications in the engineering area that used the dragonfly algorithm is offered. The utilized engineering applications are the applications in the field of mechanical engineering problems, electrical engineering problems, optimal parameters, economic load dispatch, and loss reduction. The algorithm is tested and evaluated against particle swarm optimization algorithm and firefly algorithm. To evaluate the ability of the dragonfly algorithm and other participated algorithms a set of traditional benchmarks (TF1-TF23) were utilized. Moreover, to examine the ability of the algorithm to optimize large-scale optimization problems CEC-C2019 benchmarks were utilized. A comparison is made between the algorithm and other metaheuristic techniques to show its ability to enhance various problems. The outcomes of the algorithm from the works that utilized the dragonfly algorithm previously and the outcomes of the benchmark test functions proved that in comparison with participated algorithms (GWO, PSO, and GA), the dragonfly algorithm owns an excellent performance, especially for small to intermediate applications. Moreover, the congestion facts of the technique and some future works are presented. The authors conducted this research to help other researchers who want to study the algorithm and utilize it to optimize engineering problems.

**Nomenclature**

As scientific texts use different notations for vectors, matrices, random variables, derivatives concerning vectors, Table 1 provides the convention applied in this paper.

**Table1: List of symbols used in the paper**

| Symbol | Meaning |
|--------|---------|
| f | Attraction towards food |
| s | separation |
| w | inertia weight |
| c | cohesion |
| a | alignment |
| e | distraction outwards predators |
| L1 | Lower Bound |
| Ui | Upper Bound |

## 1. Introduction

Many researchers in various areas use Swarm Intelligence (SI). The ability of natural swarm systems amazed natural scientists and biologists to study the behaviors of swarms and creatures. Swarm-based algorithms are part of the nature-inspired population-based algorithm's family. Regarding complex real-world problems, these algorithms produce good results in terms of cost, speed, and robustness [1]. Bonabeau mentioned SI as the evolving of the combined intellect of sets of modest representatives [2]. Swarm intelligence systems consist of several agents that form a population. They consist of a collection of intellectual performance of systems that are self-organized and decentralized. Collective clustering and sorting, building nests, and foraging groups of social insects are examples of SI-based techniques [3]. As discussed in [2], labor division and self-organization are two basic concepts of SI. Self-organization here means the capability of having procedures for developing agents with no support from external sources. On the other hand, the labor division indicates the implementation of numerous feasible with meek jobs by people. In SI, agents follow simple rules, and no centralized control structure exists to control the behaviors of individuals. In reality, the individual's behaviors are local and random to an extent. Artificial individuals, however, interact with each other, which produces intelligent and new actions [4]. SI recently has been applied to different problems in continuous and combinatorial optimization, robotics, telecommunications, etc., and often magnificent results were produced [5]. Lately, the researchers have proposed some new techniques. Particle Swarm Optimization (PSO) was suggested by Kennedy and Eberhart [6]. PSO is one of the first-born algorithms in the swarm intelligence field. It mimics the behaviors of a collection of fish or birds. In the exploration space, each particle is a particular agent with a location. Since inventing the PSO, its original and improved versions have been used to optimize many complex problems, for example, references [7-9]. Additionally, He et al. proposed Group Search Optimizer (GSO) [10]. GSO mimics the searching behavior of animals. Cuckoo Search (CS) algorithm imitates the process of reproduction in the cuckoo family [11]. Later in 2014, Mirjalili et al. developed Grey Wolf Optimizer (GWO) [12]. It imitates the hunting behavior of wolves. Later, about [13], Mirjalili proposed Dragonfly Optimization Algorithm (DA). DA mainly mimics the behaviors of hunting and migration of dragonflies. Harmony Search Algorithm (HA) is proposed in [14]. It mimics the process of improving music by the musician. The musician tries to provide better harmony depending on his/her experiences. Donkey and Smuggler Optimization (DSO) algorithm suggested in [15]. DSO imitates the attitudes of donkeys to select and search routes. Yazdani et al. developed another example of nature-inspired algorithms, which is called the Lion Optimization Algorithm (LOA) [16]. The LOA mimics the lion's cooperation behavior and their unique lifestyle. Based on social organization, the lions divide into residents and nomads. The residents consist of several lions that live together, and they are called pride. Nomads, on the other hand, are mostly seen in pairs and sometimes singularly. Lions may change their lifestyle from nomads to residents or vice versa. Moreover, Rahman, C., and Rashid have proposed a new Learner Performance-Based Behavior (LPB) algorithm [17]. The LPB mimics the process of accepting graduated students from high school in different colleges. Similar to other algorithms, LPB produces the initial population randomly. In the later steps, depending on the fitness of the individuals, the population is divided into several sub-populations. The optimization process starts from the sub-population that contains the best individuals, and then the next best sub-population, and so on. This procedure avoids locating into local optima and also provides a good balance between exploration and exploitation. Multi-objective versions of metaheuristic algorithms are also utilized to optimize multi-objective problems, such as reference [18].

Different researchers have used DA in numerous diverse applications and it gave satisfactory results. Until the end of working on this review paper (March 2019), almost 300 different works cited the dragonfly algorithm in different areas. It produced satisfying results in almost all applications. Additionally, the authors of this review paper published another review paper on the DA and its applications in applied science [19]. In that review paper, the authors cantered their review on the applied science area (such as image processing, machine learning, wireless, and networking). Dragonfly algorithm is used for optimizing a huge number of problems in various disciplines. One review paper cannot cover all the articles that used the DA. Thus, in this paper, DA and its engineering applications are focused on and reviewed.

The main objectives and contributions of the work are: 1) Presenting one of the newly developed metaheuristic algorithms for optimization called DA. 2) Moreover, discussing the engineering problems that utilized the DA. 3) Comparing the DA to several metaheuristic optimization algorithms including (FA, GWO, MFO, HHS). 4) Examining the ability of DA against several benchmarks and comparing the results to the (GWO, PSO, and GA). 5) Nevertheless, collecting the works that used DA in the field of engineering in one paper to help researchers who want to use the algorithm in this field.

This work first shows a short overview of the dragonfly algorithm in section two. Next, we discuss the variants of the algorithm in section three. Afterward, in section four, the authors address some of the hybridization versions related to the DA algorithm with other algorithms. In section five, the applications that were solved by the DA in the field of engineering are presented. Additionally, in section six, the DA is compared with other metaheuristics. In section seven, the advantages and disadvantages of the reviewed algorithm are presented. In section eight, the algorithm is evaluated using the traditional benchmark functions and the Congress on Evolutionary Computation set (The 100-Digit Challenge) or so-called (CEC) benchmark functions. The evaluations are then compared with the GWO, PSO, and GA. Furthermore, in section nine, a discussion and some problems that encounter the DA's operators are dealt with in conjunction with giving explanations and prospect works for enhancing the capability of DA. Lastly, the key points of this research work are established in section ten.

## 2. Dragonfly Algorithm

In the last few decades, the natural behavior of creatures has widely motivated metaheuristic optimization algorithms. Swarm intelligence is the main inspiration for metaheuristics [6, 20]. DA is a metaheuristic optimization method. It imitates the swarming attitudes of dragonflies [13].

Dragonflies are little predators. They hunt insects in nature. The main reason for the dragonflies swarming is hunting and migration; in other words, these are two phases; static and dynamic swarms, respectively. In the first phase, which is the static swarming, a set of dragonflies generate sub-swarms and search through different small areas. On the other hand, in the second phase, which is dynamic swarming, a set of dragonflies can fly in a much bigger swarm. They fly in one direction towards the most promising global optimum area [13].

In dynamic swarming, dragonflies maintain a reasonable separation and cohesion (intensification or exploitation). In static swarming, conversely, alignment is too big; cohesion is small for attacking prey (diversification or exploration).

Therefore, small cohesion and great alignment weights will be assigned to individuals once exploring the search space. However, they will be assigned to high cohesion and low alignment weights while exploiting the search space. The neighborhood radii proportionally enflamed to the iteration number for changeover between intensification and diversification. Another way for balancing intensification and diversification is tuning the swarming weights adaptively during the process of optimization. The swarming weights are; attraction motion towards food (f), separation (s), inertia weight (w), cohesion (c), alignment (a), and distraction outwards predators (e). The Following are the equations for the swarming weights:

Reynolds in [21] mentioned that Equation (1) can be used for computing separation:

$$S_i = -\sum_{j=1}^{N} X - X_j \qquad (1)$$

X signifies the current individual's position.
$X_j$ specifies the $j^{th}$ dragonfly's position in the neighborhood.
N designates the dragonflies' number in the neighboring.
S signifies the $i^{th}$ dragonfly's separation motion.

The alignment can be calculated by using Equation (2) [13].

$$A_i = \frac{\sum_{j=1}^{N} V_j}{N} \qquad (2)$$

$A_i$ specifies the motion of alignment for $i^{th}$ dragonfly.
V specifies a $j^{th}$ dragonfly's velocity in the neighborhood.

Equation (3) for calculating cohesion:

$$C_i = \frac{\sum_{j=1}^{N} X_j}{N} - X \qquad (3)$$

C specifies the $i^{th}$ dragonfly's cohesion.
N specifies the neighborhood size.
$X_j$ specifies the $j^{th}$ dragonfly's position in the neighborhood.
X specifies the present individual.

Equation (4) is for calculating attraction motion towards food:

$$F_i = X^+ - X \qquad (4)$$

$F_i$ specifies the attraction of food of the $i^{th}$ individual.
X+ specifies the food source's position.
X specifies the current individual's position.

Equation (5) is for calculating distraction outwards predator:

$$E_i = X^- + X \qquad\qquad (5)$$

$E_i$ specifies the distraction motion of the enemy for the i$^{th}$ dragonfly.
$X^-$ specifies the position of the enemy.
X specifies the dragonfly's current position.

Individuals' positions of the artificial dragonfly are updated in the exploration space utilizing vectors, namely; ΔX, which is called step vector and X, which is called position vector. ΔX in the dragonfly algorithm is equivalent to the velocity in particle swarm optimization. Updating the position of individuals in DA mainly depends on the PSO algorithm. Whereas X specifies the movement direction in dragonfly individuals. X can be computed as follows [13]:

$$\Delta X_{t+1} = (sS_i + aA_i + cC_i + fF_i + eE_i) + w\Delta X_t \quad (6)$$

s specifies the weight of separation.
$S_i$ specifies the i$^{th}$ individual separation.
a specifies the weight of alignment.
$A_i$ specifies i$^{th}$ dragonfly's alignment.
c specifies the weight of cohesion.
$C_i$ specifies i$^{th}$ dragonfly's cohesion.
f specifies the weight of food attraction.
$F_i$ specifies the i$^{th}$ individual food source.
e specifies the weight of enemies' distraction.
$E_i$ specifies the i$^{th}$ dragonfly's enemy position.
w specifies the weight of inertia.
t indicates a counter for iterations.

Once calculating the ΔX, the calculation for the X starts in this manner:

$$X_{t+1} = X_t + \Delta X_{t+1} \qquad\qquad (7)$$

$t$ specifies current iteration.

We should add a random move to the searching technique to upsurge the exploration likelihood of the entire choice space through an optimization technique. When neighboring solutions do not exist to fly over throughout the exploration space, the dragonflies would use a method of random walk or so-called Lévy flight. Here, the dragonfly's position is modified as follows:

$$X_{t+1} = X_t + Lévy(d) \times X_t \qquad\qquad (8)$$

As mentioned $t$ indicates the present iteration and $(d)$ specifies the position vector's dimension.

Reference [22] stated that although using the Lévy flight improves the performance of DA, however, it might cause very long steps. In the mentioned reference, to avoid this drawback, the Brownian motion was used in place of the Lévy flight. The motion of Brownian is another mechanism of random motion. The free liquid or gas molecules movement has inspired this. The modified DA complexity was O; the size of the population multiplied by the iteration number. The calculated complexity proved that using Brownian motion did not have an impact on the complexity time of the original DA. By using the Brownian motion, the massive jumps caused by the Lévy flight were corrected. However, occasionally sudden moves may still be required to avoid trapping into local optima. For objectives with local minima, the Brownian motion produced better solutions in a shorter time.

For the changeover between exploitation and exploration, dragonfly individuals change their weights adaptively. To adjust the flying path during the process of optimization, the neighborhood area should be enlarged, hence before the optimization process ends; the whole swarm becomes one group for converging to the global optimum.

## 3. Variants of Dragonfly Algorithm

Dragonfly algorithm has three different versions:

### 3.1. Single Objective Problems-Dragonfly Algorithm

Like most Si-based optimization algorithms, DA initially creates a solution set randomly for the optimization problem in hand. At first, the position and step vectors of individuals were assigned to arbitrary values between both upper and lower variables' bounds. Positions and step vectors are updated for all dragonflies per iteration. For updating the vectors; position and step, the dragonfly's region is selected through the Euclidean distance calculation between all the individuals. Iteratively, the individual's position updating continues until the end criterion is met.

### 3.2. Binary Dragonfly Algorithm

Since in binary search space only 0 or 1 can be assigned to the position vector, adding step vectors to the position vector cannot update the position of search agents. The transfer function produces a binary technique from a continuous SI technique. The velocity (step) values work as an input to the transfer function, and then the transfer function yields a number between (0 and 1) as result, which states the likelihood of moving the individuals and updating their position. Alike to continuous optimization, the transfer function reproduces unexpected variations in particles by significant velocity.

Equation (9) is for computing the probability of changing the positions of all dragonflies [23].

$$T(\Delta X) = \left| \frac{\Delta X}{\sqrt{\Delta X^2 + 1}} \right| \qquad (9)$$

In binary search spaces and after utilizing Equation (9), Equation (10) updates the location of the search agent. $r$ ranges between 0 and 1.

$$X_{t+1} = \begin{cases} \neg X_t & r < T(\Delta X_{t+1}) \\ X_t & r \geq T(\Delta X_{t+1}) \end{cases} \qquad (10)$$

Binary Dragonfly Algorithm (BDA) assumes that all of the individuals are in one swarm. Hence, it adaptively tunes the swarming factors such as s, f, c, a, e and w to simulate intensification and diversification.

Reference [24] used BDA for feature selection. This work proved the importance of the role of the transfer function for producing the discrete space from the continuous one and an enhanced balance concerning the phases of exploitation and exploration. The proposed work stated that Equation (9) does not provide the right balance concerning the phases of exploitation and exploration, where at the start of the optimization, the exploration rate should be higher than exploitation. Hence, to raise the BDA's performance and avoiding falling into local optima, a time-dependent Transfer Function (TF) was used.

The value of time-dependent TF linearly gets bigger as the step vector of the search agents gets more significant. Consequently, in the early steps of the algorithm, higher exploration is provided. However, as time passes the probability of exploitation increases and the probability of exploration decreases. As proved in this work, the examined TF has enhanced the performance of the BDA. The main reason for this was providing the correct balance concerning the phases of exploitation and exploration of the BDA.

The computational complexity of the BDA using the time-dependent TF is the same as the original BDA, and it is O(ISD). Where $I$ indicates the iteration number, $S$ indicates the solution number, and $D$ indicates the dimension number.

### 3.3. Multi-Objective Dragonfly Algorithm

These problems have more than one objective. The answer to this type of difficulty can be often known as a set of Pareto optimal. The finest compromise among the existing objectives present in the Pareto optimal set [25]. The Pareto optimal dominance compares two solutions in multi-objective search space [26]. DA is first provided with an archive for saving superior solutions of Pareto optimal throughout the process. The food source would come from the archive to update the position, and the rest of the procedure is similar to that of the DA.

Likewise, the Multi-Objective Particle Swarm Optimization (MOPSO) algorithm [27], for finding the best Pareto optimal front, the food source can be chosen from the minimum populous zone of the present Pareto optimal front. For including the entire solutions, a hypersphere can be defined. In each iteration, equal sub-hyper-spheres are produced by dividing the hyper-spheres. Whenever the segments are produced, for each segment a roulette-wheel technique with a specific probability is utilized for the process of selection [28].

Multi-Objective Dragonfly Algorithm (MODA) has a better likelihood of selecting the food source from a smaller amount of populous segments. Contrarily, to choose hunters from the record or archive, the equation chooses the worst or most populous hyper-sphere, so that the dragonfly individuals are discouraged to hunt about unpromising zones.

In each iteration, the archive updates regularly, and it may become full during the process of optimization. Hence, to prevent that situation there should be a technique. If as a minimum one of the habitations govern the solution, then it should not go into the records. On the other hand, if the solution controls some of the solutions of Pareto optimal, then the solution will be added to the record, and all the Pareto optimal solutions will be deleted. If the archive or record becomes full, some solutions from the most populated segments will be deleted [28]. MODA has two extra parameters, which do not exist in the DA: one of the parameters is for describing the max number of hyper-spheres, and the second one is for defining the size of the archive.

In reference [29], the authors modified the multi-objective DA. The proposed algorithm was named as Multiobjective Dragonfly Algorithm based on the Reference point (RMODA). In the examined work, a sequence of reference points was placed. Every individual in the population was set with the smallest reference point with is Euclidean distance. Every reference point corresponds to several individuals. To keep the population size, individuals are selected from the key layer to be inserted into the population. The procedure of selecting the reference points with the minimum number of correlated solutions is preferred. If the number of associated references is not zero, the point is deleted; if not, the solution owns the minimum distance to the reference point is selected by one of the key layers. This process continues until the size of the Pareto optimal solution set equals the population size. To improve the distribution ability of the proposed algorithm, an external archive was utilized. Whenever the number of individuals becomes bigger than the size of the population, the cutting technique from the Improved Strength Pareto Evolutionary Algorithm (SPEA2) [30] is utilized to keep the size of the population. The proposed algorithm was utilized to optimize the Wind-Solar-Hydro power generation system. The provided results were compared to the results of the NSGA-III [31] to solve the same problem. The results showed that the distribution of the solutions provided by RMODA, and the convergence of the RMODA were better compared to the NSGA-III.

## 4. Hybridized Versions of Dragonfly Algorithm

One of the most popular techniques to enhance the ability of metaheuristic algorithms is merging the strong properties of different algorithms. As a result, a novel algorithm will be produced based on the features of the amalgamated algorithms [32]. Some of the hybridized DA versions in other areas are discussed in [19]. The rest of the hybridized versions of the DA can be discussed in this section.

As mentioned, the exploitation and exploration phases of DA are examined through alignment, cohesion, separation, and attraction toward food and distraction against enemies. The aforementioned searching technique improves the exploration and diversity of the solutions. On the other hand, a huge number of social interactions in DA cause trapping into local optima, less accuracy, and improper balance between exploitation and exploration. Additionally, DA fails to secure the best-founded solution in the previous iteration. Consequently, the algorithm may not converge to global optima, and the exploitation ability of the algorithm decreases. To overcome these deficiencies, in reference [33], DA combined with Improved Nelder-Mead (INM) algorithm for improving the local exploration of the algorithm and avoiding falling into local optima. The INM is a two-way extensive search algorithm. It is an improved version of the Nelder-Mead (NM) unidirectional search algorithm. Throughout the searching procedure of the NM, updating the worst and best position of the population entirely depends on the position of the current individual. If the foremost global individual found so far is not the global optimal solution, for other particles it impulsively traps into local optima. To conquer this behavior of the NM, a two-way extensive searching technique dependent on centroid search is designed in the INM algorithm. In the INM, the algorithm's centroid is moving continuously towards the global best with precise weight and outwards the worst position. Consequently, at the end of the process, the centroid is very close to the global best position and far away from the worst one. The best properties of the INM algorithm include the good performance of optimization and protecting the best and worst position and centroid which improves the ability of local search of the algorithm and avoiding trapping into local optima. The mentioned properties of the INM can avoid the weak points of DA. Hence, they are combined to produce a better algorithm. Two techniques were added to the DA to enhance its ability for optimization: in the first technique, a memory matrix was added to save the best-founded solutions in the previous iterations; in the second technique, the saved solutions were then utilized as initial input to the INM algorithm for more exploitation. The produced INM-DA hybrid (INMDA) can be divided into two steps, in the first step, the DA is utilized to explore the solution space. It provided a necessary exploration ability to the artificial dragonflies to find the global optimum. The second step utilized the INM algorithm to find the worst and the best point and calculating the population centroid. The key feature of the INM was that the centroid of the population was utilized to update the position. Hence, the chance of trapping into local optima is reduced. For high-dimensional problems, the produced results proved that the examined work performed better compared to the DA and Memory-Based Hybrid Dragonfly Algorithm (MHDA) and that they are not a good choice for solving high dimensional problems as they rapidly run into a dimensional curse. The high performance of the proposed work came from the improved ability of both exploitation and exploration of reverse learning techniques.

In reference [34], the DA's strength is combined with Artificial Bee Colony (ABC). The ABC algorithm mimics the behaviors of honeybees. It consists of three main phases, each phase conforms to behaviors of a specific class of honeybees. The classes are onlookers, employed, and scout bees. Each class has a part in the foraging procedure, collecting information about the quality of the food and assessing the source of the food based on the provided information. The quality of a solution is examined depending on the quantity of nectar in the source of the food. Information about the location of food sources is collected by the employed bees. The collected information by the employed bees is then returned to the beehive. The onlookers work on the collected information by the employed bees and decide on the best food source. The mentioned procedure continuous for several iterations, and when no improvements occurred in the standard of food source for some employed bees, the source of food is abandoned, and the employed bee turns to a scout bee. A new stochastic search is started by the scout bee to discover new food sources. The ABC owns a good ability to find local optima through the employed and onlooker phases. It mainly depends on selecting the individuals that enhance the local search. Onlooker and employed bees phases are recognized as the local search operators. However, the scout bees phase is recognized as the global search, which reduces the speed of convergence throughout the searching procedure. The position updating procedure in the DA algorithm utilizes the *Levy Flight*, which causes large moves and makes finding the global optima difficult. In reference [28], the DA and ABC were combined to eliminate the convergence speed problem and falling into local optima by providing a better steadiness concerning local and global search constituents of the contributed techniques. The proposed hybrid algorithm is called the Hybrid ABC/DA approach (HAD). It consists of three phases: the Onlooker Bee, the Dragonfly Bee, and the Modified Scout Bee. Two main enhancements were made in the proposed hybrid algorithm; first, modifying the scout bee phase to enhance the diversity of searching procedure and improving the efficiency of global search in the ABC. The second enhancement was to replace the employed bee phase in the ABC algorithm with operators from the DA. The enhanced operator was called Dragonfly Bee Phase. The main idea behind the HAD was to combine the exploration of ABC with the exploration and exploitation of DA. The exploitation of the original DA is great, and the exploration of the ABC is good. Hence, combining them in the proposed hybrid algorithm provides very good performance and convergence. Providing two exploration steps to the HAD algorithm widens the searching space of the algorithm which improves the diversity of individuals. Thus, the HAD algorithm searches and exploits a larger area to find the global optima. The proposed hybrid algorithm was examined against several benchmarks and problems. The results were compared to the original DA and ABC, and a set of other algorithms. It was shown that the proposed HAD algorithm provided better results and performance in most of the cases.

Reference [35] proposed an adaptive DA to optimize frame structures. In this article, Coulomb Force Search Strategy (CFSS) combined with and a new hybrid algorithm produced. The proposed algorithm is named as Coulomb Force Search Strategy-Based Dragonfly Algorithm (CFSS-DA). CFSS is a new searching technique that is utilized for fastening the convergence. The CFSS inspired by the power among charges in an electromagnetic field. One of the main characteristics of CFSS is that when some charges repulse, the other charges attract one another, which is alike the exploration and exploitation phases. In the CFSS, the position and velocity of individuals are updated using data on many optimal solutions. Here, two dragonflies are assumed to be attracting each other. The exploratory constant parameter (k) is one of the essential parameters in the Coulomb force search strategy. This work examined the utilization of adapted value (k) in the course of the searching procedure of the dragonfly algorithm. The dragonflies were encouraged for searching in the search space with giant steps at the beginning of the process and small steps at the end of the process. The above-mentioned adaptive strategy improved the convergence of the algorithm. Hence, it produced an optimal result in a short time evaluated against the standard algorithm, then the compared results to the DA and BDA from the proposed technique proved this. The proposed algorithm is used to optimize the front axle of an automobile. In the examined problem, the front axle beam was selected. The outcomes substantiated that the convergence speed of the CFSS-DA compared with the BDA and DA is much better and that the proposed hybrid algorithm can be utilized to optimize engineering problems and produce reliable optimal solutions in a few iterations.

## 5. Applications of DA in Engineering

The ability of DA encourages numerous academics to apply it to optimize different applications in various areas. In the following subsections, we discuss applications of the dragonfly algorithm in Engineering and Physics.

### 5.1. Mechanical Engineering

Network configuration is the practice of altering the position of open or closed switches to make changes in the distribution network's topological structure. In [36], a new reconfiguration schema was developed to reduce the net deviation among the nominal voltage value, and the node voltages using a dragonfly optimization algorithm. Dragonfly Optimization Algorithm Based Reconfiguration Method (DORM) enhanced the Voltage Profile (VP) by the Net Voltage Deviation (NVD) minimization. The proposed technique was examined without making any thermal violations. It has also kept the radial structure. In this study, the results obtained using DORM compared to some other nature-inspired algorithms for solving configuration problems, such as PSO [37], GA [38], and BBO [39]. According to the study, the obtained results proved that the DORM provided better configuration through minimizing NVD and providing a good VP. To share the loads with the conventional power plant, distribute generation units utilized. The mentioned units are also used to give power to the loads individually. Wind Turbine (WT), Photovoltaic (PV), Gas Turbine (GT), Micro-Turbine (MT), and Storage Battery (SB) are the most typical distributed generation units in this type of application.

In reference [40], a novel optimal scheme of a different Hybrid Power Generation System (HPGS) is generated. The introduced design consisted of a combination of PV, WT, GT, and SB. Natural gas distribution networks are utilized to fuel the GT of the system. To find the optimal design of the proposed work, two metaheuristic techniques; DA [13] and GWO [12] were examined. The system considered different weather conditions. Both metaheuristic algorithms in this work are used for minimizing the annual cost and entire emission functions for the system. It concluded that the DA produced better results in respect of the total yearly cost comparing to GWO. In contrast, regarding the system pollution, the GWO technique produced better results than the DA technique.

Perforated plates are part of many industrial applications in recent years. Perforated plate cutouts are mostly used to decrease the structure weight or to build a point of exit and entry. Cutouts in the plates can change the geometry of the plate, which leads to severe local stresses or called stress concentration throughout the cutouts. This can cause a reduction in strength and premature failure in structures. Therefore, knowing useful parameters to reduce stress concentration in various structures is crucial. In reference [41], DA is used to optimize the involved parameters in analyzing the stress of the perforated orthotropic plates. The aim was to achieve the minimum stress value nearby the quasi-triangular cutout positioned in a boundless orthotropic license plate. Dissemination of stress computed employing the suggested technique established on the analytical solution of Lekhnitskii. The variables that were designed using the proposed technique included load angle, material properties, bluntness, fiber angle, and cut-out orientation angle. The outcomes were compared to the PSO and Genetic Algorithm (GA). The factors of the stated algorithms are shown in Table 2. The outcomes evidenced that the regular of best values of stress produced via the DA was smaller compared to other algorithms. It concluded that the values of both average and standard deviation for the DA were smaller than the GA and PSO [42]. The comparison of these techniques proved that DA showed the excellent performance to solve the problem mentioned above, and it operated more steadily. It was also determined that the high exploration and exploitation rates in the DA made the algorithm perform better. Moreover, DA converged much earlier (18th iteration), whereas PSO and GA converged in the iterations 95th and 146th, respectively. Additionally, depending on the results, it was observed that the most significant levels of stress in all cutout bluntness or w happened on 45°-load angle.

**Table 2: Parameter Settings of the Algorithms [41]**

| DA | GWO | PSO | GA |
|---|---|---|---|
| Population size = 100 | Population size = 100 | Population size = 100 | Population size = 100 |
| Max. No. Of iterations = 1000 | Max. No. Of Iterations = 1000 | Max. No. Of Iterations = 1000 | Max. No. Of Iterations = 1000 |
| Random values = $r_1 = r_2 = [0, 1]$ | | Cognitive component = $c_1 = 2$ | Probability of crossover (Pc) = 0.8 |
| Separation weight (s′) = 0.1 | | Social component = $c_2 = 2$ | Probability of Mutation (Pm) = 0.03 |
| Alignment weight (a′) = 0.1 | | $\omega = \dfrac{0.1}{\left|\frac{1}{2}\frac{c}{2}\frac{\sqrt{|c^2-4c|}}{2}\right|}$, $c = c_1 + c_2$ | |
| Cohesion weight (c′) = 0.7 | | | ncrossover = 2*round(npop *Pc/2) |
| Food factor (f′) = 1 | | | nmututation = npop*Pm |
| Enemy factor (e′) = 1 | | | |
| Inertia factor ($\delta+ = 0.9$-0.2) | | | |
| Constant ($\zeta$) = 1.5 | | | |

The robust non-linear link concerning the array factor and the array's elements that marks the Concentric Circular Antenna Array (CCAA) problem synthesis is challenging. A high Maximum Side Lobe Level (MSL) is a problem of CCAAs. Reference [43] used DA to design CCAA in a way that was able to get low side lobes. A Sub-Structured Neural Network (SSANN) was used instead of a single Artificial Neural Network (ANN), which improved the forecast accuracy of the effectiveness of requalification sub-ANNs and the engine working process. The proposed work aimed at observing and exploring the effectiveness of the DA technique. Moreover, in this work, four different CCAA design cases were used to study DA efficiency. Then, the results evaluated against approaches like BBO [44], SOS [45], SQP [44], CSO [46],

OGSA [47], EP [48], and FA [49]. The proposed work utilized two three-ring designs; CCAA with 4-, 6-, 8- plus 8-, 10-, 12-, besides two cases well-thought-out for each model: CCAA without, and with the center component. For each scheme test, the space between neighboring elements in every ring was fixed to 0.55, 0.606, and 0.75 from the center to the outermost ring. The outcomes of the DA were evaluated against the techniques in the literature of the work, and a uniform array was utilized. The outcomes showed that the DA had better performance for the mentioned problem, and it was competitive with other methods for decreasing MSL.

In reference [50], automatic generation control of an interconnected two-area multi-source hydrothermal power system is considered. The performance of the scrutinized system was evaluated and planned with Proportional-Integral (PI), Proportional Integral Derivative (PID), and 2 Degrees Of Freedom PID (2DOF PID). The DA was used to optimize the controller gains. It concluded that the DA provided superior results compared to classical methods. Furthermore, the 2DOF PID controller optimized by DA produced smaller values for overshoot (OS), settling time (ST), and undershoot (US). Moreover, smaller values of the objective function are provided compared to the 2DOF PID controller optimized by Differential Evolution.

Optimization can significantly affect the process of grinding by improving the quality of products and reduce operational costs and time of production. Optimizing the grinding process is a challenging process in the engineering field because of the complexity and nonlinearity of the process. In reference [51], multi-objective DA is used for obtaining solutions of non-dominated Pareto optimal. In this work, an experimental example in [52] was used. Then, the outcomes were evaluated against the outcomes of an experimental model using NSGA-II in [52]. The solutions of Pareto optimal produced via MODA conquered the attained solutions via the NSGA-II. The outcomes showed that MODA accomplished better compared to the NSGA-II in resolving the multi-objective mathematical model of the grinding process, the reason for this superiority was due to the MODA's efficient operators evaluated against the simple operators of NSGA-II (crossover and mutation). The solutions produced by MODA improved surface roughness significantly and reduced the costs and the total grinding time. The results proved that all the objectives were optimized by MODA simultaneously through the algorithm's efficient operators. MODA used 30 individuals and 1000 iterations to examine the mathematical model of the tri-objective of the grinding process. On the other hand, NSGA-II utilized 100 chromosomes and 1000 iterations that caused 100,000 function evaluations. The results proved that the MODA's computational cost was much lower than the NSGA-II's.

Reference [53] used MODA for optimizing the performance of Switched Reluctance Motor (SRM) powered by autonomous stacked Proton Exchange Membrane Fuel Cells (PEMFC). MODA is used to produce the best sets of driving circuit turn-on/off angles. As mentioned, the best sets produced via DA could improve the savings in energy and increase the performance of isolated PEMFC-SRM. Dragonfly's ability in developing the initial stochastic population and the good exploitation and exploration of DA were the reasons aimed at the superiority of the algorithm for solving this problem. Furthermore, DA provides a high uniformly disseminated Pareto optimal set of solutions in problems of multi-objective [54].

### 5.2. Electrical Engineering

A new technique for designing, modeling, and optimizing a uniform serpentine meander based on MEMS switch incorporating beam puncture effect was discovered in [55]. A new analytical model was suggested, which aimed at pull-in voltage in this research work. An optimization technique was introduced for finding the best configuration of the switch to accomplish the least possible pull-in voltage. Here, the analytical model was used as an objective function. For this purpose, the author utilized several great evolutionary optimization methods for achieving the best measurements with less cost computationally and more simplicity. The conducted techniques included PSO, DE, a hybrid PSO with Differential Evolution (DEPSO), DA, WOA, and Human Behavior Based PSO (HBPSO). A comparison among the applied algorithms showed that the DA had the best minimum pull-in voltage with the smallest errors. The parameter settings for DA in the proposed work were: dimension = 8, search agents = 50, alignment weight, separation weight, and cohesion weight were random between -0.2 and 0.2, food attraction weight was random, and enemy distraction weight was a value between -0.1 and 1. The results showed that the DA performance was the best to minimize pull-in voltage with minimum errors.

In the power transmission system, the stability of voltage is a significant concern due to inconsistency between demand and power generation. Reference [56] utilized the Eigenvalue Decomposition (EVD) method and DA in partitioned Y-admittance matrix to identify weak buses for implementing the compensators of reactive power. In this work, DA is used to enhance the static VAR compensator's size and cost. Regarding the objective function, line flows, voltage deviation, and reactive power limit was examined as the design constraints. The results proved that the proposed technique maximized the cost of static VAR compensator and the cost of installation with the loading condition. In addition, the voltage deviation and the actual power loss in the DA were much smaller compared to the PSO. Moreover, the DA could show its superiority in reducing real power loss for the IEEE 30 bus system compared to the other algorithms, additionally, DA converged earlier.

The Atomic Generation Control (AGC) problem was examined in reference [57] by using DA. In this work, the DA optimized the control parameters, for example, scaling factors of fuzzy logic and PID gains. The criterion of Integral of Time Multiplied Absolute Error (ITAE) was used to minimize the settling time with a minimized peak overshoot. The ITAE is employed for optimizing the scaling factor and PID gains controller. The addressed control strategy was examined through two equal non-reheat thermal interrelated power system areas. The work stretched to two hydrothermal power system areas joined via a High Voltage Direct Current (HVDC) transmission link and an AC tie line. To deal with non-linearity, the Generation Rate Constraint (GRC) effect is counted. The results proved that in terms of lowest damping oscillations, settling time, peak undershoots, and overshoot in the interrelated three-area power system through GRC non-linearity, the proposed metaheuristic algorithm based fuzzy PID controller provided superior results evaluated against further control methods. The results proved that the DA as an optimization technique produced a better optimum solution of AGC for non-linear and linear interconnected power systems' frequency regulation. Furthermore, the combined fuzzy PID controller proposed in this work proved its superiority over the fuzzy logic and optimized PID controller.

### 5.3. Optimal Parameters

Reference [58] optimized the factors in the examining stress of perforated orthotropic plates. In this work, the DA was utilized to compute the stress distribution based on the analytical solution of Lekhnitskii. Fiber angle, load angle, orientation cutout, bluntness, and material properties are included in the study design variables. The results obtained from the dragonfly algorithm in this work were evaluated against the results of GA [59] and PSO [6]. The results proved that in comparison to the PSO and GA, the DA converged earlier. Besides, avoiding local optimum and producing better results proved the DA's supremacy compared with the other two algorithms. The DA also produced smaller average values of optimum stress compared to the other algorithms. Furthermore, by using the DA, a standard deviation closer to zero was produced, which was smaller than the ones produced using the PSO and GA.

Providing reliable and continuous supply to customers is a critical ambition of utility and meets the expectations of power balance and the loss of transmission when the generators operate within a specified limit. For achieving this purpose, the value of emission and the fuel cost ought to be as insignificant as conceivable. The allowed deviation in feasible tolerance and fuel cost is named as Emission Constrained Economic Dispatch (ECED) problem. Reference [60] used DA for finding an explanation for the problem of ECED. In this work, the value of emission and the fuel cost alongside the quadratic function was treated as a problem with many objectives. To convert the problem to a single objective, the price penalty factor technique is used. The consequences of penalty factors, such as Min-Min, Min-Max, Max-Max, Max-Min emission value of different gas exhalations, and price penalty factors are mentioned in this work. As the results in this work showed that using "Min-Max" as the price penalty factor produced less fuel cost compared to the other penalty factors, however, increasing ECED fuel cost by 17% could reduce emission by almost 23% in comparison with the price penalty factor of "Min-Max". The author mentioned that nowadays having a small amount of ECED fuel cost to operate a thermal power plant with a "Min-Max" price penalty creates contamination in the environment and causes premature death in humans leaving near the thermal power plant.

Reference [61] introduced a new technique to participate in online engine calibration and to control increasing the performance of the engine and decreasing gas emission of the greenhouse. For this purpose, the mentioned reference used a robust model centered on a multi-objective genetic algorithm or NSGA-II, multi-objective dragonfly algorithm, fuzzy dependent on inference system, and Sub-Structural Neural Network (SSANN). Throttle angle, injection angle, engine rpm, and injection time were used as the inputs for SSANN. The Fuel Flow (FF), CO, torque, and NOx were used as outputs. Initially, the data from GT-POWER was used to train SSANN. Based on various engine speeds, 15 working points were selected randomly to examine the accuracy of SSANN. Linear regression was utilized for assessing the linear relationship between the measured and predicted outputs. For this problem, MODA converged earlier (at the 40th generation), and it had better Inverse Generation Distance (IGD). However, NSGA-II converged after the 80th generation. In addition, it was discovered that with increasing the number of iterations MODA showed better convergence. It was because of the use of the food/enemy selection technique in the MODA.

About [62], the vibrant strength of the Hybrid Energy Distributed Power System (HEDPS) is considered. The HEDPS was subject to wind power and load variations. A controller with Three Degrees Of Freedom (3-DOF) Proportional-Integral-Derivative (PID) was implemented and designed in the HEDPS to balance frequency fluctuations and power after the perturbation. Unlike the Single-Degree-Of-Freedom (1-DOF) controller, the 3-DOF controllers own the ability of an outstanding set-point tracking, and it produced superior regulations for the input disturbance. DA is used for optimizing the factors of 3-DOF PID controllers. Also, Integral Time Absolute Error (ITAE) was used as an optimizer to optimize the 3-DOF controller gains. The achieved outcomes were evaluated against the outcomes of other popular metaheuristic algorithms, such as Zeigler-Nichols (ZN). The isolated, interconnected modes of hybrid energy and distributed power system are implemented for assessing the proposed controller's performance. For qualitative assessment, the convergence of DA was evaluated against the other participated algorithms. The outcomes demonstrated that the dragonfly algorithm established the value of global optimum by a quicker rate and that a lesser minimum value for the fitness function was generated compared to the other participated algorithms. For this work, all the algorithms generated the optimal global point between 60 to 70 generations, which gave the choice of having 100 iterations. Furthermore, the results concluded that the DA outpaced the other stated algorithms with regards to faster convergence and the value of minimum fitness.

### 5.4. Economic Load Dispatch

A wind integrated system with the valve-point effect was considered in [63]. DA used to overcome the problem of Economic Load Dispatch (ELD) along with the valve-point effect. The Weibull distribution function was used to model the stochastic nature of wind. Furthermore, a closed integral function was used to analyze the overestimation/underestimation cost. In the proposed work, the optimization technique started by generating a set of random solutions for the assumed problem. The dragonfly's vectors (position and step) were randomly initialized within the upper and lower bounds of generators. The outcomes exhibited that the DA successfully resolved the power system of the economic dispatch of the wind thermal integrated system. Two cases and the IEEE-30 bus system were implemented for calculating the performance. The problem of non-convex economic dispatch was solved in the first case. The obtained results from this case compared to a Sequential Quadratic Programming Particle Swarm Optimization (SQP-PSO) technique. ELD with wind power penetration was solved using DA in the second case. Moreover, the performance of the work in case 2 compared to SQP-PSO [64]. In both cases, 1200MW is considered as a load demand. The results showed that the DA found a global optimum solution and it was remarkably unrestricted from locating into local optima.

In reference [65], the dragonfly algorithm was applied to improve a novel technique to resolve economic dispatch incorporating solar energy. In carrying out the economic dispatch, the mentioned reference considered prohibited operating zone and valve-point loading constraints. The beta distribution function is applied for modeling the solar energy system and the objective function. The output is predicted to include four diverse periods. Various loading circumstances are considered for each. The proposed work addressed that compared to other optimization methods dragonfly algorithm gave a low cost, minimum power loss, and converges in the minimum running time. It is concluded that more power could be generated if the availability of the sun was abundant in the chosen location. Moreover, in the case of using the produced system power correctly, the economy will maximize, and the system loss will minimize. The proposed work considered three different cases. In case 1, the system used for testing consisted of six generators and 1263 MW. The results from the first case study compared to the most recent optimization methods. The results proved that the DA was the best regarding the convergence time, the smallest objective function, power loss, and evaluations. Similarly, in terms of generations, cost, and transmission loss, DA was the best. Concerning case 2, the number of used generators was 15, and 2630 MW was considered. Here, the total cost generation for the DA was minimum. In case 3, the 86 bus test system was utilized in south Indian. It consisted of 7 generators, 131 lines, and 86 buses. This case considered ramp rate constraint, transmission loss, down reserve constraints, and up the reserve. Here, the obtained results proved that the optimal cost of DA was much smaller than the completive algorithms.

### 5.5. Loss Reduction

The research work [66] is based on the BDA. A new technique for wrapper selection was proposed in the research work. The suggested technique aimed at diminishing the number of characteristics concerning the standard feature set and obtain better accuracy in classification at the same time. The K-Nearest Neighborhood (KNN) classifier is applied to test the selected subset of the feature. The subset of feature selection is a problem of multi-objective. Problems of multi-objective study two diverse goals. The proposed work aimed at maximizing the accuracy of classification, and diminishing the features. Equation (11) shows the objective function. The proposed approach was evaluated against 18 UCI datasets. A comparison was made between the proposed technique and similar techniques that used GA, and PSO. The comparison was concerned with the accuracy of a classification and the number of carefully chosen attributes. The outcomes proved that BDA had a superior ability in examining the space of features and choosing the features with more information for the task of classification.

$$Fitness = \alpha y_R(D) + \beta \frac{|R|}{|C|} \qquad (11)$$

$y_R(D)$ shows the rate of error of the classification used.
|R| represents the selected subset's cardinality.
|C| signifies the whole number of characteristics included with the dataset.
$\alpha$ And $\beta$ signify factors representing the classification importance and length of the subset, respectively.
$\alpha \in [0, 1]$ and $\beta = (1 - \alpha)$, the author adopted these from [67].

Reference [68] solved a nearly zero-energy-building design problem. A comparison was made in terms of performance among seven multi-objective algorithms. In the utmost of the cases, the attained solutions were enhanced by increasing the generation number. Each algorithm ran 20 times with moderately raising the evaluation number. The optimization results in most running cases proved that the results of MODA were uncompetitive. In terms of contribution and running time, MODA was not competitive, and it was slow. According to this work, MODA did not have any outstanding features.

Power loss, electric distribution system's maximum loadability, and Voltage Stability Margin (VSM) are greatly affected by inadequate reactive power generation. To solve these problems, about [69], optimal concurrent as well as multiple separate installations of Distributed Generation (DG), and capacitor were examined. For this work, minimizing the total of Reactive Power Loss (QL) counted as the primary objective, and DA was used to optimize the problem. Standard 33-bus distribution systems were utilized to test the methodology proposed in this work. The proposed work handled different capacitor and DG installation cases. The results of the proposed work compared to weight improved particle swarm optimization or WIPSO technique. The results proved that the primary behavior of DA for updating the individual's position provided an enhanced QL reduction compared to the other methods. The results also showed a better convergence rate by producing fitter solutions in 15 to 20 iterations.

## 6. A Comparison between Dragonfly Algorithm and Other Algorithms

Reference [70] addressed an assignment of court cases that has an impact on enhancing the effectiveness of the jurisdictional structure. The effectiveness of the jurisdictional structure extremely relies on punctuality and operating the court cases efficiently. In the proposed work, Mixed-Integer Linear Programming (MILP) was utilized to solve the problem of assigning cases in the justice court. The objective function of this issue was assigning N cases to M groups. Each group might cope with the cases altogether. However, because of the requirement of the cases, personal potentiality, and other assigned cases, the necessary time for each group to solve the same case was not the same. To find the best solution for the proposed work, DA and the Firefly Algorithm (FA) was utilized [47]. Two problems were assessed in a uniform distribution. In the form P1: Lower bound (Ll) = (1, 30), Upper bound (Ui) = (1, 90), efficacy rate ($\mu_i$) = (1, 90), and P2: Lower bound (Ll) = (1, 60), Upper bound (Ui) = (1, 90), efficacy rate ($\mu_i$) = (1, 90). The outcomes exhibited that for finding the best solution the DA required less time and an average percentage deviation to maximize efficacy compared to the firefly algorithm. The outcomes proved that in 50 cases and three-justice groups aiming at trial parameters: P1 (50:3, 4, 5) and P2 (50: 3, 4, 5), the DA was greater compared to FA.

In [71], GWO, DA, and Moth-Flame Optimization (MFO) algorithms were assessed for optimizing the best sitting of the capacitor in several Radial Distribution Systems (RDSs). The loss sensitivity factor was examined for discovering the candidate buses. The authors considered 33-, 69-, and 118-bus RDSs to prove the efficiency and effectiveness of the addressed optimization technique. This study aimed at minimizing the total cost with voltage profile improvement and power loss. The outcomes were evaluated against the outcomes of the PSO for showing the advantage of the utilized methods. The GWO-, DA-, and MFO-based techniques produced better outcomes about the PSO-based technique concerning several iterations and the convergence speed for the addressed issue. Furthermore, for the 69-bus distribution system case, DA-, GWO-, and MFA-based optimization exhibited an enhanced convergence level. Additionally, GWO, DA, and MFO were assessed using statistical tests. The results showed that GWO, DA, and MFO had an acceptable Root-Mean-Square Error (RMSE).

Reference [72] introduced a novel binary multi verse optimization algorithm. In the article, the authors compared a new algorithm to some other binary optimization algorithms, including BDA. BDA ranked as the second-best algorithm, among others, this was because of the excellent stability between DA's exploitation and exploration phases. Furthermore, the sudden changes in the variables provided a quick convergence to the BDA.

Reference [73] compared DA with the Harris Hawks Optimization Algorithm (HHS). The algorithms were utilized to enhance the multi-layer perceptron's performance, which was used to analyze the stability of two-layered soil. The work compared the accuracy and computational time of the algorithms. Mean Absolute Error (MAE), the area under the receiving operating characteristic curve (AUC), and Mean Square Error (MSE) were utilized for evaluating the predictive models' performance. In general, both algorithms helped to improve the applicability accuracy of the MLP. However, the DA reached the lowest error within 500 iterations, whereas the HHS needed 1000 iterations for the same task. Hence, the DA provided a better convergence comparing to the HHS for the problem mentioned above.

## 7. Advantages and Disadvantages of Using DA

The good sides of the DA that attract the researchers to utilize it in many applications include:
   1.   The implementation of the algorithm is not difficult.

2. It suits very well with engineering problems and even other problems in different areas. This behavior of the algorithm attracts many researchers to use it for optimizing various problems in different fields.
3. Its selection procedure avoids locating into local optima and searching around a non-promising area.
4. It has few parameters to tune.
5. It has a good convergence time.

On the other hand, the algorithm has some drawbacks, such as:
1. It does not have an internal memory, which causes trapping into local optima. However, this problem is fixed in reference [74].
2. The high exploitation rate of the algorithm leads the DA to trap into local optima.
3. Utilizing the Levy flight mechanism caused overflowing in the search area and an interruption in the stochastic flights.

## 8. Results and Evaluations

The results of the applications in the literature showed that the dragonfly algorithm is suitable to optimize various applications in the engineering field. The provided outcomes proved the superiority of the algorithm. Here, to demonstrate the ability of the DA, it is evaluated against the traditional benchmark functions. To examine the ability of the algorithm and its performance, three groups of traditional benchmark functions were utilized with various characteristics in the original work. The groups of the traditional test functions consist of three groups, which are unimodal (F1-F7), multi-modal (F8-F13), and composite test functions (F14-F23). Unimodal test functions examine the exploitation and convergence ability of the algorithm. As their name shows, this group of benchmark functions has a single optimum. On the other hand, multi-modal benchmarks have a single global optimum, and more than one local optimum. The algorithm should have the ability to avoid all the local optima and go toward the global optima. Hence, multi-modal benchmarks examine the exploration ability of the algorithm and the ability of the algorithm in avoiding many local optima. The composite benchmarks are biased, combined, shifted, and rotated versions of the unimodal and multi-modal benchmarks [17]. They examine the obstacles in the real search spaces by providing a large number of local optima and diverse shapes to different regions. This type of test function examines the balance between exploitation and exploration of the algorithm. The authors tested the DA, GWO, and PSO to examine the algorithms against the test functions. The results of test functions are shown in Tables 2 and 3.

Moreover, to further evaluate the algorithm, it was examined on the IEEE Congress of Evolutionary Computation Benchmark Test Functions or CEC-2019, also known as "the 100-digit challenge" [75]. CEC-2019 benchmarks examine the ability of the algorithm to optimize large-scale optimization problems. The CEC01 to CEC03 own different dimensions as shown in Table 4. However, the rest of the CEC benchmarks are 10-dimensional problems in the [-100, 100] ranges, and they are rotated and shifted test functions. All the CEC benchmarks have global optima towards 1. To evaluate all the algorithms 100 search agents over 1000 iterations over 30 independent runs were utilized for all the participated algorithms and test functions. The results of all the benchmarks are shown in Tables 3 and 5. The results in bold indicate the superior results.

It can be seen in Table 3, the DA provided better results in 3 unimodal test functions out of 7. Hence, in optimizing unimodal test functions the DA outperformed the GWO, PSO, and GA. The results of the unimodal test functions are evident that the DA has outstanding exploitation and better convergence speed compared to the other participated algorithms. Nevertheless, the results from the references mentioned in the paper are another piece of evidence for the speed of convergence of the DA. Reference [47] utilized the DA and FA for optimizing the same problem. The results showed the high convergence of the DA.

Table 3: Classical Benchmark Results of DA, GWO, PSO, and GA

| Test Functions | Measurements | DA | GWO | PSO | GA |
|---|---|---|---|---|---|
| TF1 | Mean | **0** | 1.02972476885246e-177 | 1.00885499599309e-57 | 0.00115137781562392 |
| | Std. | **0** | **0** | 3.00874945514207e-57 | 0.00117730979184216 |
| | Time (Sec.) | 4427.518096 | 20.641365 | **11.368635** | 897.760482 |
| TF2 | Mean | 0.346105250457283 | **2.68628277893028e-99** | 1.43990230057245e-21 | 0.00472661522708145 |
| | Std. | 1.82441635559704 | **5.80256327360805e-99** | 3.39557465863375e-21 | 0.00403239542723310 |
| | Time (Sec.) | 3346.272200 | 24.371195 | **13.382481** | 173.993546 |
| TF3 | Mean | **0** | 2.68197560527247e-86 | 1.68861236134112e-18 | 17.1472265934958 |
| | Std. | **0** | 7.27762974655797e-86 | 5.48465114022047e-18 | 12.2898587657973 |
| | Time (Sec.) | 3240.021810 | 46.182206 | **42.488767** | 159.290370 |
| TF4 | Mean | **0** | 1.22497881456461e-57 | 9.99789072283524e-17 | 0.291636037814873 |
| | Std. | **0** | 2.15924903796066e-57 | 1.95740963479449e-16 | 0.0893850074302745 |
| | Time (Sec.) | 3130.959784 | 20.685333 | **11.413006** | 78.483755 |
| TF5 | Mean | 125.910484817663 | 5.97036252298473 | **3.44770889218702** | 18.6154267152628 |
| | Std. | 640.972765080855 | 0.620546420114094 | **1.70882531582746** | 23.2769550298819 |
| | Time (Sec.) | 3850.276794 | 24.916651 | **18.117383** | 92.735804 |
| TF6 | Mean | 1.34292399356453 | 4.43642866073549e-07 | **0** | 0.00162926113184969 |
| | Std. | 0.752756708459943 | 1.24192058527820e-07 | **0** | 0.00176098997870077 |
| | Time (Sec.) | 3811.746878 | 21.643090 | **15.897804** | 99.301974 |
| TF7 | Mean | 0.000633011935049757 | **8.06568003572986e-05** | 0.00166365483908592 | 0.00739102668757681 |
| | Std. | 0.000896536807762407 | **7.62436276091404e-05** | 0.000688786606476328 | 0.00482154555572441 1 |
| | Time (Sec.) | 3802.443433 | 27.678768 | **22.298551** | 77.130858 |
| TF8 | Mean | -3140.99378304017 | -3039.99432880358 | -2746.74174287925 | **-3741.60266890918** |
| | Std. | 392.935454710813 | 335.770432754438 | 318.016650960271 | **204.313660817165** |
| | Time (Sec.) | 4479.760791 | 23.528453 | **19.729532** | 142.192202 |
| TF9 | Mean | 0.135097099077087 | **0** | 1.89042220847802 | 0.000672751769171024 |
| | Std. | 0.726281615039210 | **0** | 1.17870035792479 | 0.000658014717969960 |
| | Time (Sec.) | 3349.950311 | 20.065488 | **15.729011** | 75.932717 |
| TF10 | Mean | **8.88178419700125e-16** | 4.44089209850063e-15 | 4.44089209850063e-15 | 0.0129635567438434 |
| | Std. | **0** | **0** | **0** | 0.00865984802947136 |
| | Time (Sec.) | 3137.754139 | 32.066467 | **19.150031** | 142.430190 |
| TF11 | Mean | **0** | 0.0148131151617185 | 0.130078492564645 | 0.060199643096524 |
| | Std. | **0** | 0.0309176914861047 | 0.0926149652075663 | 0.0275406722596615 |
| | Time (Sec.) | 3250.038645 | 25.495103 | **23.885916** | 114.662938 |
| TF12 | Mean | 0.269342484156197 | 9.21079314060305e-08 | **4.71486124195438e-32** | 9.37245789759370e-05 |
| | Std. | 0.344192758720291 | 3.33881013489254e-08 | **1.76746048719996e-34** | 0.000235863174207910 |
| | Time (Sec.) | 3882.885198 | **48.455417** | 57.892857 | 109.184386 |
| TF13 | Mean | 0.693499718050741 | 5.13830672973745e-07 | **1.34978380439567e-32** | 0.000311832171596598 |
| | Std. | 0.277012042085844 | 1.78986623626440e-07 | **5.56739851370242e-48** | 0.000796848182889176 |
| | Time (Sec.) | 3824.597920 | **59.435809** | 63.901939 | 109.246994 |
| TF14 | Mean | **0.998003837821488** | 1.32868739097539 | 1.16367501486549 | **0.998003837800378** |
| | Std. | 1.46216765100100e-10 | 0.752071664102933 | 0.376784985644918 | 2.15871205008169e-11 |
| | Time (Sec.) | 3178.074526 | 175.376344 | **138.331474** | 195.261368 |
| TF15 | Mean | 0.00154211186710959 | **0.000338011460252839** | 0.005815436762252451 | 0.009810407463000232 |
| | Std. | 0.000555025989291821 | **0.000167180666774048** | 0.000268239587689150 | 0.000408252211563814 |
| | Time (Sec.) | 4006.598768 | **17.247217** | 17.329560 | 77.046801 |
| TF16 | Mean | -1.03162845346044 | -1.03162845139082 | -1.03162845348988 | -1.03162842656173 |
| | Std. | 1.61221884746401e-10 | 2.35250806658275e-09 | **6.77521542490044e-16** | 1.47491527943430e-07 |
| | Time (Sec.) | 2926.022886 | 13.741340 | **5.792700** | 75.462114 |
| TF17 | Mean | **0.397887357729738** | 0.397887590554324 | **0.397887357729738** | **0.397887357729738** |
| | Std. | 3.24338745434428e-16 | 3.55849349665776e-07 | **0** | **0** |
| | Time (Sec.) | 3076.341415 | 12.927428 | **7.551330** | 63.943990 |
| TF18 | Mean | **2.99999999999992** | 3.00000068941874 | **2.99999999999992** | 3.00000802124552 |
| | Std. | **2.32224382025056e-15** | 5.26140732330268e-07 | **1.27488253802344e-15** | 4.37983852341747e-05 |
| | Time (Sec.) | 3261.595454 | 12.034492 | **9.832012** | 88.187258 |
| TF19 | Mean | **-3.86227373322268** | -3.86223673347316 | **-3.86278214782075** | -3.75971052872133 |
| | Std. | 0.00136979970619921 | 0.00195411179304832 | **2.71008616996018e-15** | 0.267266881488563 |
| | Time (Sec.) | 3305.615855 | **23.114955** | 30.854334 | 130.544395 |
| **TF20** | Mean | -3.23240911060208 | -3.25541611249950 | **-3.28236413114427** | -3.27830816793432 |
| | Std. | 0.105022108234188 | 0.0643846493780909 | **0.0570048884719485** | 0.0582313584337596 |
| | Time (Sec.) | 5108.207511 | 31.980021 | **17.609688** | 117.322374 |
| **TF21** | Mean | **-9.52508869710175** | -9.30788454034324 | -8.12311378128791 | -8.72945318519225 |
| | Std. | **1.84659874404817** | 1.92203630274357 | 2.52888434537084 | 2.90197010675982 |
| | Time (Sec.) | 3382.691839 | 58.069403 | **45.103100** | 136.935151 |
| **TF22** | Mean | -9.77362416583620 | **-10.4026771080218** | -9.87552878983335 | -10.1730804844161 |
| | Std. | 1.64720106908786 | **0.000149398330570923** | 1.60928404809095 | 1.21798768605337 |
| | Time (Sec.) | 3432.630731 | 61.315575 | **51.764072** | 118.754221 |
| **TF23** | Mean | -9.96571473264067 | **-10.5362033520525** | -10.3577177141906 | -10.3097071129624 |
| | Std. | 2.00266702824654 | **0.000103330271308728** | 0.978736553880816 | 1.22280380037701 |
| | Time (Sec.) | 3401.605909 | **73.988416** | 75.872207 | 149.648881 |

The results of the test functions of the multi-modal showed the great exploration level of the dragonfly algorithm that aids in discovering the exploration space. Generally, the results provided by the DA and PSO for the multi-modal test functions were better compared to the GWO and GA. Each of DA and PSO provided better results in two multi-modal test functions while each of GWO and GA provided better results in only one multi-modal test function. The results of multi-modal test functions show the great exploration ability of the DA and PSO and their ability in avoiding a large number of local optima and go towards the global optima.

Moreover, the DA outperformed the GWO, PSO, and GA to optimize the composite test function. The DA provided better results in 6 composite test functions, PSO in 5, GWO in 4, and GA in 3 composite test functions. The results for this group of benchmarks proved that the DA has a superior balance between the phases of exploration and exploitation compared to the GWO, PSO, and GA.

Concerning the processing time, it was proved that the PSO is the fastest algorithm among the participated algorithms to produce the results and that the DA is the slowest algorithm among them.

Furthermore, in the original work, the CEC benchmark functions were not used to evaluate the DA. Hence, in this paper, the test functions of CEC-C2019 are used to assess the DA further. This group of test functions utilizes an annual optimization competition. Professor Suganthan and his colleges improved these benchmark functions to optimize single objective problems [75].

**Table 4**: CEC-C2019 Benchmark Functions- 100-Digit Challenge [75]

| Function Name | Functions | Dimension | Range | $f_{min}$ |
|---|---|---|---|---|
| CEC-C01 | Storn's Chebyshev Polynomial Fitting Problem | 9 | [-8192, 8192] | 1 |
| CEC-C02 | Inverse Hilbert Matrix Problem | 16 | [-16384, 16384] | 1 |
| CEC-C03 | Lennard-Jones Minimum Energy Cluster | 18 | [-4, 4] | 1 |
| CEC-C04 | Rastrigin's Function | 10 | [-100, 100] | 1 |
| CEC-C05 | Grienwank's Function | 10 | [-100, 100] | 1 |
| CEC-C06 | Weiersrass Function | 10 | [-100, 100] | 1 |
| CEC-C07 | Modified Schwefel's Function | 10 | [-100, 100] | 1 |
| CEC-C08 | Expanded Schaffer's F6 Function | 10 | [-100, 100] | 1 |
| CEC-C09 | Happy Cat Function | 10 | [-100, 100] | 1 |
| CEC-C10 | Ackley Function | 10 | [-100, 100] | 1 |

For this group of test functions, DA is compared to the GWO, PSO, and GA. The default parameter settings were not changed during the optimization. For this evaluation, the authors used 1000 iterations and 100 agents. As shown in Table 5, the GWO outperformed the rest of the participated algorithms, and that the DA showed poor performance compared to the other algorithms.

Similarly, Regarding the processing time, PSO again proved that it is the fastest algorithm among the participated algorithms to produce the results and that the GWO is the second-best algorithm, and the DA is the slowest algorithm in providing results.

**Table 5: The IEEE CEC-2019 Benchmark Results for DA, GWO, PSO, and GA**

| Test Function | Measurements | DA | GWO | PSO | GA |
|---|---|---|---|---|---|
| CEC01 | Mean | 7344680255.96508 | **20195906.9114461** | 117383140915.921 | 4124767802.83302 |
| | Std. | 14179795998.1417 | **63763970.4725385** | 85001697002.1108 | 2412923816.29294 |
| | Time (Sec.) | 5845.602840 | **957.919637** | 1124.206827 | 976.073305 |
| CEC02 | Mean | 19.2578777353509 | **17.3431778874724** | 5917.21564947965 | 17.5982242432266 |
| | Std. | 0.761158449891241 | **9.17145135912341e-05** | 1887.26018240798 | 0.561414181151120 |
| | Time (Sec.) | 5480.316183 | 21.757981 | **13.243870** | 173.930683 |
| CEC03 | Mean | **12.7024204195666** | 12.7024042189533 | 12.7024042179563 | 12.7024042214617 |
| | Std. | 3.84195766940490e-05 | 1.13206887070580e-09 | **3.61344822661357e-15** | 7.18124380259203e-09 |
| | Time (Sec.) | 5780.184521 | 32.965731 | **19.407994** | 194.519015 |
| CEC04 | Mean | 802.818385908163 | 39.4605871732290 | **8.35771547651643** | 142.476006906475 |
| | Std. | 953.246774635009 | **17.4319699629934** | 33.70064011826850 | 49.0142274425374 |
| | Time (Sec.) | 4940.283421 | **21.983706** | 63.907078 | 190.371677 |
| CEC05 | Mean | 1.89296693657886 | 1.28523009771205 | 1.11835647185838 | **1.09472805293053** |
| | Std. | 0.487787543643066 | 0.186783054802480 | **0.0946656630897843** | 0.0962002842697390 |
| | Time (Sec.) | 12557.704416 | 22.742999 | **15.573494** | 186.515107 |
| CEC06 | Mean | 8.66021328746622 | 9.98231955852709 | 6.44047475449302 | **4.39133549128376** |
| | Std. | 1.24430384243394 | **0.666708446314702** | 1.84881250989753 | 0.926390878021783 |
| | Time (Sec.) | 5140.436560 | **316.270789** | 362.208678 | 358.184318 |
| CEC07 | Mean | 459.680771220159 | 290.835349473841 | **102.880891701219** | 215.904023663734 |
| | Std. | 187.743428228146 | 194.569888322255 | **120.590801032533** | 144.785286959863 |
| | Time (Sec.) | 5019.963538 | **23.110244** | 23.110244 | 96.631880 |
| CEC08 | Mean | 5.43805609875719 | **4.17900130888240** | 4.66981668835183 | 4.60223045775513 |
| | Std. | **0.569983847703406** | 0.870621944305451 | 0.674616524865225 | 0.734086337081485 |
| | Time (Sec.) | 4482.366124 | 22.587019 | **16.005946** | 93.901634 |
| CEC09 | Mean | 15.8273076560544 | 3.87517009427681 | **2.74182540592406** | 3.03699188466418 |
| | Std. | 19.6442833016762 | 0.840865518421628 | **0.00181347752237683** | 0.320573048634389 |
| | Time (Sec.) | 3869.030640 | 20.432201 | **13.141358** | 99.479410 |
| CEC10 | Mean | 20.2315204551779 | **19.7638080850226** | 20.0349876978763 | 19.8371218596057 |
| | Std. | 0.128897713654758 | 3.15287575519112 | **0.0587052740290958** | 4.44534723269840 |
| | Time (Sec.) | 4415.185673 | 27.221155 | **19.781788** | 95.262277 |

## 9. Discussion and Future Works

DA is modest and can be easily applied. For exploring the search space, allocate little weight of cohesion and great weight of alignment to individuals. Contrarily, for exploiting the exploration space, assign individuals to high cohesion and low alignment weights. Another way for balancing exploitation and exploration is adaptively adjusting the swarming weights, such as s, a, e, c, w, and f throughout the process of optimization. To make a transition concerning the exploration and exploitation, neighborhood radii enlarged proportionally to the number of iterations could be applied. It usually provides reasonable results for small to medium-scale problems. However, for large-scale optimization problems, more affords are required, and it causes an increase in convergence time and a reduction in performance, which may cause falling into local optima.

With growing the complexity of optimizing real-world problems, computing demands are hard to be satisfied with the single version of optimization algorithms. One obstacle that may occur during using the DA is that updating position in this algorithm is not so much correlated with the algorithm's population centroid in the preceding generations. Consequently, the produced solutions have low accuracy, and premature convergence to local optima may occur. Additionally, it may cause it difficult to find the global optimal solution. Furthermore, as mentioned earlier, distraction, cohesion, alignment, and separation in the direction of enemy sources with desirability in the direction of the sources of food mainly determine the exploration and exploitation of the DA. This searching technique maximizes solution diversity and makes the capability of exploration of the DA stronger to some extent. Nonetheless, the performance reduces with a large number of exploitation and exploration operators because they enlarge the convergence time, which causes trapping into local optima.

Similar to other metaheuristic algorithms, DA has several strong points, as well as some weak points. It owns powerful optimization capability. The DA has few parameters for adjusting. Most of the time, it can keep a reasonable convergence rate to the global optima. DA is one of the new algorithms. However, as discussed in the literature, it has been utilized for optimizing an enormous number of applications. The straightforwardness of DA is one of the main reasons for its contributions to various applications. Also, choosing the individuals from the record, the worst hypersphere avoids the DA from discovering the non-promised zones. Another advantage is that DA has few parameters for tuning. Similarly, over other optimization algorithms, the algorithm converges earlier, is more stable, and more straightforwardly can be hybridized with diverse algorithms.

Alternatively, for complex optimization problems, as examined in [33], one of the restrictions of the DA is that it easily traps into local optima, and it has a slow convergence speed. Internal memory does not exist in the DA, which is a reason for early arrival at local optimums. This overcame in [74] through emerging a new Memory-Based Hybrid Dragonfly Algorithm (MHDA). Additionally, as presented earlier, DA uses Levy flight as a search process when the neighborhood does not exist. Nevertheless, the giant steps of the Levy flight mechanism caused an interruption. The original work used a step control mechanism to prevent overflowing. However, this distorts the characteristics of the swarm, also, it is a reason for falling into local optima. Hence, utilizing other searching techniques instead of the Levy flight and compared the results of the various methods is highly recommended. Moreover, using an adaptive step instead of the original stochastic step will help in harmonizing phases of exploration and exploitation and enhancing the DA performance. The position updating technique is another way to prevent trapping into local optima. Using the population's centroid technique, as discussed in [76] can reduce the probability of locating into local optima.

Furthermore, after assessing the algorithm in the above section, it was noted that the ability of the DA for balancing between the phases of exploration and exploitation is low; this was because the algorithm has a great exploration level. This great level of the search in the initial phases of the course of optimization is decent, though, in the last iterations of the algorithm, it ought to be diminished, and the exploitation level ought to be improved. For binary dragonfly algorithms, for example, using a time-dependent transfer function can increase the balance between both phases of the DA; exploitation, and exploration. Hence, at the beginning of the optimization, the exploration level is great. The exploration level gradually decreases during the process and the exploitation level increases. The mentioned technique will provide a better performance and it prevents trapping into local optima. Tuning parameters automatically improves the performance of different algorithms. Moreover, it improves the stability between the two phases and the variety of the population [77]. On the other hand, the outcomes of the traditional benchmark function of the unimodal, and the produced outcomes of the majority of the literature works displayed that the dragonfly has a good convergence. The greater convergence of the algorithm makes it outperform most of the mentioned algorithms in the previous works in dealing with small to medium problem sizes.

Generally, the results from the previous section and the applications in the literature proved that the DA has a significant level of exploration and exploitation. The reason for this is the DA's static swarming behavior, which enlarges the exploration's level, and increases the probability of trapping into local optima for simple problems. Additionally, enlarging the number of iterations enlarges the exploitation degree, and enhances the accuracy of the global optimum solution.

However, hybridizing the algorithm with other techniques will give power to the algorithm to overcome the bottlenecks. As discussed, some hybrid versions of DA were proposed to overcome the weakness of this algorithm. For example, MHDA was examined for overcoming the shortage that may cause premature convergence to local optima.

Moreover, reference [78] utilized Gauss chaotic map to adjust variables. The outcomes exhibited that the hybridized algorithm concerning stability quality, the speed of convergence, classification performance, and the number of selected features provided better results. Although the DA and the hybridized DAs provided some good results for several problems of complex optimization, yet, more or fewer disadvantages were found. In DA, high exploration and exploitation acquire through desirability on the way to food and diversion on the road to enemies. The correlation of updating the position in the DA with the population centroid from the preceding generation is not high. Thus, it may solve with that traps into local optima, low accuracy, and struggles to find the global optima. Hence, finding a new technique for updating the position of individuals is highly recommended. Another research area that will improve the algorithm is finding suitable stability concerning phases of exploration and exploitation. Proper stability concerning exploration and exploitation will circumvent DA from falling into the local optima. Besides, merging new searching methods with the DA is highly recommended to researchers. Moreover, tuning parameters dynamically during the practice of optimization will have significant guidance on enhancing the exploitation and exploration balance of the algorithm. Moreover, the DA is highly recommended to be utilized in other works, such as [79] and [80].

## 10. Conclusions

This paper reviewed one of the new metaheuristic algorithms. The various types of the algorithm, including the merging versions with other techniques, were discussed. In addition, most of the optimization problems in engineering and physics that used DA were discussed. From the reviewed works, the authors discovered that DA is one of the practical techniques in the area. The simplicity of the algorithm was one of the reasons that encouraged the researchers to use the algorithm to optimize the problems at hand. Moreover, the accuracy and convergence speed of the algorithm are other reasons. For instance, in general, for small to medium problems, the algorithm provided good results. However, similar to other algorithms, for some problems (especially complex problems) DA cannot produce reasonable results. The exploration of the algorithm is high, which may cause trapping into local optima, mainly for complex problems. Moreover, the produced results from the test functions (F20-F23) proved that the results of the DA are better compared to the GWO, PSO, and GA. However, the results of the CEC-2019 test function showed that the other participated algorithms outperform the DA in optimizing large-scale optimization problems. Finally, reviewing the DA and its applications proved that the DA can be utilized successfully to optimize almost all the problems in the real world.

As an extension of this work, the authors are willing to find a technique for providing a decent balance between the exploration and exploitation phases of DA. Likewise, the representation of the DA can be assessed and evaluated against other well-known and competitive algorithms, such as Donkey and Smuggler Optimization Algorithm [15], WOA-BAT Optimization Algorithm [81], Fitness Dependent Optimizer [82], Modified Grey Wolf Optimizer [83], etc.

**Acknowledgments:** The authors would like to thank the University of Kurdistan Hewler for providing facilities for this research work.

**Funding:** This study is not funded.

**Conflict of Interest:** The authors declare that they have no conflict of interest.

**Ethical Approval:** This article does not contain any studies with human participants or animals performed by any of the authors.